\documentclass[letterpaper, 10 pt, conference]{ieeeconf}  
\usepackage{cite}
\usepackage{hyperref}
\usepackage{balance}
\usepackage{amsmath,amssymb,amsfonts}
\usepackage{pifont}  
\usepackage{caption}  
\usepackage{graphicx}
\usepackage{float}

\usepackage{algorithm}
\usepackage{algorithmic}

\usepackage{multirow}
\usepackage{multicol}
\usepackage{array}
\usepackage{booktabs}
\newcolumntype{P}[1]{>{\centering\arraybackslash}p{#1}}

\usepackage{color}

\IEEEoverridecommandlockouts                              

\title{\LARGE \bf
Efficient Object Rearrangement via Multi-view Fusion
}

\author{Dehao Huang$^{1, 2}$, Chao Tang$^{1, 2}$, Hong Zhang$^{1, 2}$ \emph{Fellow, IEEE}
\thanks{$^{1}$Shenzhen Key Laboratory of Robotics and Computer Vision, Southern University of Science and Technology, Shenzhen, China.}%
\thanks{$^{2}$Department of Electronic and Electrical Engineering, Southern University of Science and Technology, Shenzhen, China.}
}

\begin{document}

\maketitle
\thispagestyle{empty}
\pagestyle{empty}

\begin{abstract}

The prospect of assistive robots aiding in object organization has always been compelling. In an image-goal setting, the robot rearranges the current scene to match the single image captured from the goal scene. The key to an image-goal rearrangement system is estimating the desired placement pose of each object based on the single goal image and observations from the current scene. In order to establish sufficient associations for accurate estimation, the system should observe an object from a viewpoint similar to that in the goal image. Existing image-goal rearrangement systems, due to their reliance on a fixed viewpoint for perception, often require redundant manipulations to randomly adjust an object's pose for a better perspective. Addressing this inefficiency, we introduce a novel object rearrangement system that employs multi-view fusion. By observing the current scene from multiple viewpoints before manipulating objects, our approach can estimate a more accurate pose without redundant manipulation times. A standard visual localization pipeline at the object level is developed to capitalize on the advantages of multi-view observations. Simulation results demonstrate that the efficiency of our system outperforms existing single-view systems. The effectiveness of our system is further validated in a physical experiment.

\end{abstract}

\section{INTRODUCTION}
\begin{figure}[t]
    \centering
    \centerline{\includegraphics[width=0.5\textwidth, clip=true, trim=0.0in 0.0in 0.1in 0.0in]{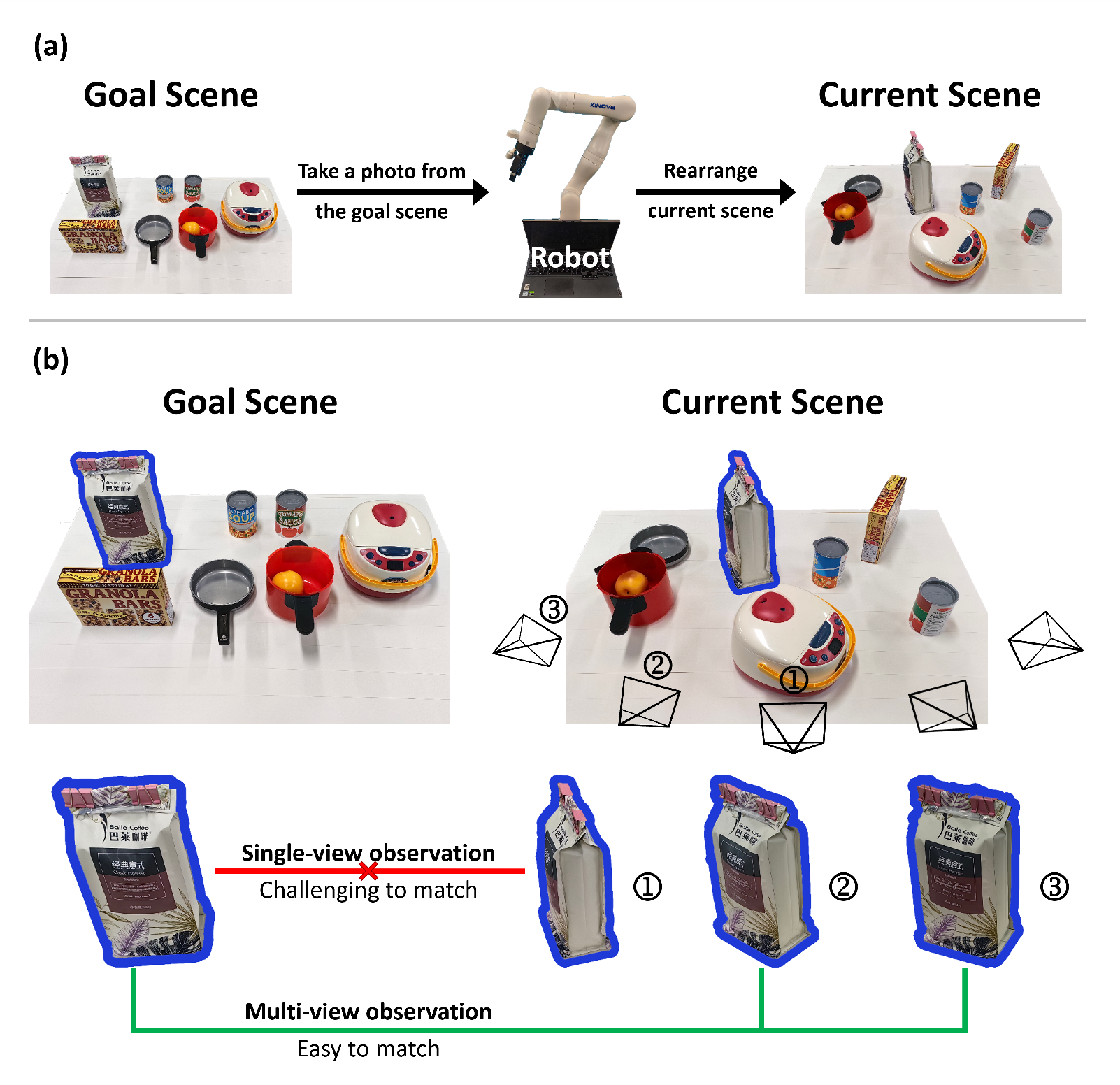}}
    \caption{(a) Our image-goal rearrangement setting. Taking a single image of the goal scene, the system rearranges the current scene to the goal scene. (b) Comparison of single-view and multi-view in object matching. In the presented scene, the object poses challenges for matching under the single-view premise. }
    \label{main}
    \vspace*{-0.28in}
\end{figure}

Object rearrangement to achieve a specific configuration has real-world applications, ranging from streamlining cluttered desks to organizing kitchenware. Beyond its practical significance, object rearrangement has been proposed as a canonical task for embodied AI\cite{batra_rearrangement_2020},  since it rigorously challenges a robot's capabilities in perception, planning, and actuation. In this work, we address the image-goal rearrangement task\cite{qureshi_nerp_2021, tang_selective_nodate, goyal_ifor_2022} in which the robot rearranges objects based on a single RGB image of the goal scene, as shown in Fig. \ref{main}(a).


Remarkable advancements have been observed in research on image-goal rearrangement. Qureshi implemented the first system for repositioning unknown objects\cite{qureshi_nerp_2021}, while Tang enhanced capabilities for more cluttered scenes and Goyal takes into account the orientation of objects\cite{tang_selective_nodate, goyal_ifor_2022}, respectively. Their perception modules estimate the desired pose of objects by matching the current single observation image with the goal image from the same viewpoint, in a single-view setting. From a single viewpoint, only a partial view of an object can be seen. If the observation from this viewpoint differs significantly from the observation in the goal image, accurately estimating the desired placement pose of the object becomes challenging due to the lack of shared features, as shown in Fig. \ref{main}(b). For precise pose estimation, existing systems require additional manipulations to randomly adjust the object's pose for a better perspective. However, increasing the number of manipulations reduces the system's efficiency.

To address this challenge, we introduce a perception module that utilizes multi-view observations, which are easily obtained before manipulating objects. Without the cost of manipulating objects, a more complete observation of an object can be used for accurate pose estimation in a multi-view setting compared with the single-view setting, as shown in Fig. \ref{main}(b). To fully utilize the multi-view observation, our approach leverages a standard visual localization pipeline for object pose estimation. We construct a hierarchical database with each object region from each viewpoint observation serving as a database item. Given the goal image, each object is first retrieved to identify the similar object regions within the database. Subsequently, pixel-level correspondences between the goal object and the retrieved object region are established through local matching. Finally, the relative pose for the planning module between the goal scene and the current scene can be solved using the PnP method\cite{lepetit_epnp_2009}.

\section{RELATED WORKS}
\noindent\textbf{Visual Object Rearrangement.} Traditional TAMP-based rearrangement systems\cite{garrett_online_2020, krontiris_efficiently_2016, krontiris_dealing_2015, wang_uniform_2021, labbe_monte-carlo_2020, wang_efficient_2022, wang_lazy_2022} link perception and planning modules sequentially and assume a priori knowledge of the objects or the scene. Their focus has been on tackling non-monotone rearrangement planning problems in complex scenarios, such as those with numerous objects and minimal buffer space. Regarding the perception module, some works\cite{wang_uniform_2021, wang_lazy_2022, wang_efficient_2022} handle only simple objects (e.g., cylinders), while others\cite{garrett_online_2020} make assumptions about known object poses. Subsequently, several systems\cite{liu_structformer_2022, liu_structdiffusion_2022, kapelyukh_dall-e-bot_2023, qureshi_nerp_2021, goyal_ifor_2022, tang_selective_nodate} introduced their task-specific perception modules and utilized existing learning-based planning modules\cite{sundermeyer_contact-graspnet_2021, tang_task-oriented_2023, tang_graspgpt_2023, danielczuk_object_2021}, significantly enhancing their ability to rearrange unknown objects. Our system belongs to the latter category and concentrates on developing a perception module.


Existing perception modules for image-goal object rearrangement are designed based on the setting of matching a single-view observation image with the goal image. For instance, NeRP\cite{qureshi_nerp_2021} correlates the point clouds of objects in the two images to construct a scene graph, predicting the subsequent object to manipulate and its goal position. IFOR\cite{goyal_ifor_2022} proposes an object-level optical flow algorithm to obtain pixel-level correspondences for the same object between the two images. These correspondences are combined with the object point cloud to predict its goal pose using the ICP algorithm\cite{segal_generalized-icp_2009}. These existing methods are inefficient due to their confinement to a single-view setting, while our approach utilizes multi-view observations before manipulating objects. This improves the accuracy of object pose estimation, subsequently reducing the number of times the object needs to be manipulated.

\vspace{0.15\baselineskip}

\noindent\textbf{One-shot Object Pose Estimation.} Beyond the applications of the visual localization at the scene level, recent works\cite{castro_posematcher_2023, he_onepose_2022, sun_onepose_2022, liu_gen6d_2022, wen_bundlesdf_2023, shugurov_osop_2022} apply the visual localization pipeline at the object level in the context of one-shot object pose estimation. Drawing inspiration from these previous works, our perception module estimates the relative object pose required for the rearrangement task based on a standard visual localization pipeline. Specifically, some works\cite{sun_onepose_2022, he_onepose_2022, castro_posematcher_2023} rely on Structure From Motion (SFM) to model a single object based on multi-view images with labeled object poses. Concurrently, they develop direct 2D-3D matching algorithms and estimate the pose of the object with PnP. Some other works\cite{liu_gen6d_2022, shugurov_osop_2022} utilize rendering or refinement techniques instead of 2D-3D matching. Due to their reliance on complete object modeling or specialized networks, these works are confined to the object scenario with only one object. In contrast, we utilize robust 2D-2D matching and introduce a hierarchical database to improve retrieval accuracy for the case of multiple objects.

\begin{figure*}[htbp]
    \centerline{\includegraphics[width=1.0\textwidth, clip=true,trim=0.1in 0.1in 0.1in 0in]{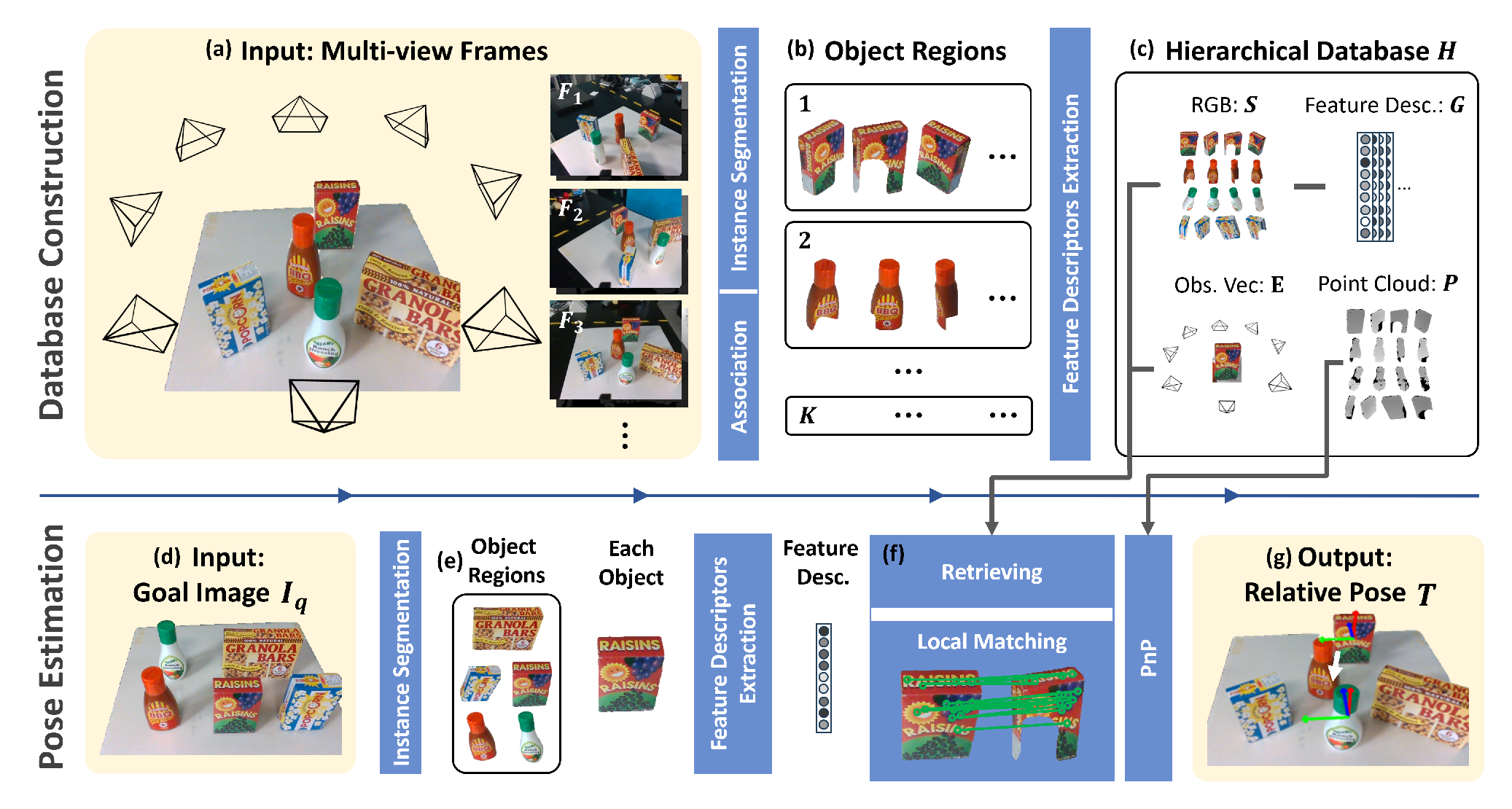}}
    \caption{Overview of the perception module. During the database construction phase, the current scene is observed from predefined viewpoints to obtain \textbf{(a)} multi-view frames $\left\{F_i\right\}$.  All frames $\left\{F_i\right\}$ are subject to instance segmentation and object-level association to obtain hierarchical \textbf{(b)} object regions. For each object region, the RGB segmentation $S$, the feature descriptor $G$, the observation vector, and the point cloud $P$ are extracted. All object regions are organized into the \textbf{(c)} hierarchical database $H$. During the pose estimation phase, the \textbf{(d)} goal image $I_q$ of the goal scene is inputted. The \textbf{(e)} goal object regions are segmented first. For each goal object region, using the extracted feature descriptors, similar object region candidates are retrieved from the database $H$. These candidate object regions are sequentially matched locally with the goal object region to obtain the \textbf{(f)} correspondences. Finally, the \textbf{(g)} relative pose $T$ of this goal object region between the goal scene and the current scene is calculated by solving the PnP problem using the correspondences.} 
    \label{pipeline}
    \vspace*{-0.20in}
\end{figure*}

\section{METHOD}
Given a single-view RGB image $I_{g}$ of the goal scene $s^*$, our system aims to rearrange the objects from the current initial scene $s$ to match the configuration of the goal scene $s^*$. Our rearrangement system consists of two core modules: a perception module and a planning module. The perception module is based on a standard visual localization pipeline. A hierarchical database $H$ is constructed using the multi-view frames $\left\{F_i\right\}$ of current scene $s$, where $i$ is the frame index. Each frame $F_i$ contains an RGB image $I_i$, a depth image $D_i$ and a camera viewpoint pose $\xi_{i} \in \mathbb{S E}(3)$. Leveraging this database, the relative poses of objects $\left\{T_j\right\}_{j=1}^K$ between $s^*$ and $s$ are calculated, where $K$ is the number of objects. In the planning module, the desired placement pose of an object is obtained by applying its $T_j$ to the grasp pose. Every object is rearranged by picking it with the grasping pose and placing it with the placement pose. This module plans to manipulate objects sequentially, aiming to avoid collisions with the environment. The system operates until all objects are confirmed to be correctly positioned. The focus of our work is on the perception module while the planning module, which has its own challenges, is implemented with an existing technique.


\subsection{Perception Module}


We design a visual localization pipeline to fully exploit multi-view information, subsequently aiming to estimate accurate object poses without manipulating objects. First, a hierarchical database $H$ is constructed for the current scene $s$. $H$ contains information that we have extracted from the original observations and can be efficiently retrieved. Then, $H$ is used to estimate the poses $\left\{T_j\right\}_{j=1}^K$ of the objects in the goal image $I_{g}$. Calculating the $T_{j}$ of an object in $I_{g}$ depends on establishing its 2D-3D correspondences with the database. The 2D-3D correspondences are derived by integrating 2D-2D correspondences from local matching in RGB images with depth information from the depth image. Below, we describe these two phases in detail.

\vspace{0.15\baselineskip}
\noindent\textbf{Database construction using multi-view frames.} 
The procedure for constructing the database $H$ is illustrated in the upper segment of Fig. \ref{pipeline}. First, we manually define a series of viewpoints $\left\{V_i\right\}$ that are spatially distributed around the center of the scene $s$. The frames $\left\{F_i\right\}$ are observed from the viewpoints $\left\{V_i\right\}$, corresponding to the RGB images $\left\{I_i\right\}$, depth images $\left\{D_i\right\}$, and viewpoint pose $\left\{\xi_i\right\}$, as shown in Fig. \ref{pipeline}(a). To model each object separately, each frame $F_i$ is subject to an instance segmentation method to obtain object regions $\left\{o_k^i\right\}$, where $k$ is the index of the object region in this frame. Every object region $o_k^i$ comprises the RGB segmentation $S_k^i$, the corresponding viewpoint pose $\xi_k^i = \xi_i$, and the point cloud $P_k^i \in \mathbb{R}^{3 \times n}$ in the world coordinate projected from the depth segmentation, where $n$ is the number of points. To perform segmentation, we combine Grounding-DINO\cite{liu_grounding_2023} and Segment-Anything\cite{kirillov_segment_2023} (SAM). By employing a predefined text prompt labeled ``objects", Grounding-DINO detects the bounding boxes of all objects. SAM then segments each object region using each bounding box as a prompt.

After frame-by-frame segmentation, the object regions between frames lack association. An object in the scene corresponds to object regions across multiple frames. This association information can reduce redundant local matching during the pose estimation phase. Hence, we construct this based on the point cloud $P_k^i$. K-means clustering is applied to the centers of the point clouds for all object regions. Each cluster is considered an object instance. All object regions are reorganized according to their corresponding object instances, e.g., a list of object regions $\left\{o\right\}_m$ corresponds to an object instance, where $m$ is the object instance id and $m \in [1, K]$, as shown in Fig. \ref{pipeline}(b)

In parallel with constructing the object-level associations, we extract a feature descriptor $G \in \mathbb{R}^d$ and an observation vector $E \in \mathbb{R}^3$ for each object region, where $d$ is the dimension of the descriptor. Based on the feature descriptor $G$, a goal object region can quickly retrieve similar object regions within the database $H$ during the pose estimation phase. The retrieval of object regions is expected to utilize global information to identify a matching object while recognizing the exact region relies on local details. We employ MixVPR\cite{ali-bey_mixvpr_2023} as our feature descriptor extractor, a holistic aggregation technique that uses both the global and local features of an image. A notable consideration is the inherent scale variability when observing an object in a scene from distinct viewpoints. Due to our scale normalization preprocessing, which involves padding and resizing before calculating $G$, our module exhibits stronger applicability in the case of multiple objects compared to OnePose, which relies on comprehensive mapping. The procedure to extract the feature descriptor is shown in Eq. \ref{global_description}. Turning to $E$, the observation vector of an object region defines the perspective from which the object region is observed for its corresponding object instance. Similar to the object-level associations, $E$ is used to avoid redundant local matching during the pose estimation phase. $E$ is described by the unit vector that originates from the central point of the point cloud $P$ and extends toward the viewpoint pose $\xi$.  This vector is computed using Eq. \ref{obs_pose}, where the viewpoint pose $\xi$ is transformed into a transformation matrix of the form $\mathbb{R}^{4 \times 4}$.

\begin{equation}
\label{global_description}
G = \text{MixVPR}(\text{pad\_and\_resize}(S))
\end{equation}

\begin{equation}
\label{obs_pose}
E=\frac{\xi[:3, 3]-\text{mean}(P)}{\left\|\xi[:3, 3]-\text{mean}(P)\right\|}
\end{equation}

In summary, every object region $o$ in the constructed hierarchical database $H$ comprises four components for the pose estimation: the RGB image segmentation $S$, the feature descriptor $G$, the point cloud $P$, and the observation vector $E$, as shown in Fig. \ref{pipeline}(c). These object regions are stored as enumerated $K$ object instance lists based on object-level associations. This database $H$ can be concisely described by the following equation:
\begin{align}
\label{database}
H &= \left\{ 
    \begin{aligned}
    &instance \, 1:  \left[ \left[S, G, P, E \right]_1^1, \ldots \right], \\
    &instance \, 2:  \left[ \left[S, G, P, E \right]_1^2, \ldots \right], \\
    &\ldots \\
    &instance \, K:  \left[ \left[S, G, P, E \right]_1^K, \ldots \right] 
    \end{aligned}
\right\}
\end{align}

\vspace{0.15\baselineskip}
\noindent\textbf{Pose estimation based on database.} Upon the construction of the database $H$, we are interested in estimating the required poses $\left\{T_j\right\}_{j=1}^K$ from a single-view RGB image $I_q$ of the goal scene $s^*$. First, object regions $\left\{o_j\right\}_{j=1}^K$ in $I_q$ are segmented using the previously mentioned instance segmentation method, as shown in Fig. \ref{pipeline}(e). Each goal object region $o_j$ contains the RGB segmentation $S_j$ and the viewpoint pose $\xi_j$. A brute-force method could compute the pose by leveraging our database $H$ constructed based on multi-view observations: each goal object region $o_j$ performs local matching with every object region in the database until sufficient matches are reached. However, this method is extremely inefficient and entirely dependent on the local matching method. In contrast, we employ a visual localization pipeline that first retrieves candidates and then performs local matching. The extracted feature descriptor $G$, observation vector $E$, and object-level association information within the database are fully utilized to enhance efficiency. The procedure is illustrated in the lower segment of Fig. \ref{pipeline}. 

In the retrieval phase, the goal is to identify similar object regions from the database $H$ as potential candidates $\left\{o_{cand}\right\}$ for each goal object region $o_j$. To achieve this, the feature descriptor $G_j$ of $o_j$ is derived, as outlined in Eq. \ref{global_description}. A similarity ranking is established by taking the dot product of $G_K$ with the feature descriptors of all object regions in $H$. We do not directly select the top-ranked object regions as candidates because such an approach overlooks the object-level association relationships and spatial relationships among object regions in $H$, which could result in the generation of redundant candidates and decreased efficiency. Assuming that all candidate object regions represent the same object instance, we first select a unique candidate object instance $u$ that appears most frequently among the top-ranked object regions. Subsequent to this, all object regions of the unique candidate instance $H\left[ u \right]$ are ranked as candidate object regions based on the original similarity. In addition to the candidate selection mechanism, we also employ an effective candidate traversal mechanism. During the local matching phase, if a candidate is rejected, it often indicates that the wrong orientation of the object was retrieved. Therefore, candidates that are close to the discarded one are deleted. Using the observation vector $E_{cand}$, the similarity between two object regions is calculated as the angular distance in spherical coordinates, as given by Eq. \ref{obs_pose_distance}. 
\begin{align}
&E_1 = \left[x_1, y_1, z_1\right], E_2 = \left[x_2, y_2, z_2\right] \\
\label{obs_pose_distance}
\text{d}(E_1, E_2)& = \text{norm}(\left[
    \begin{aligned}
    \text{arctan}(\frac{y_1}{x_1}) &- \text{arctan}(\frac{y_2}{x_2}), \\
    \text{arccos}(z_1) &- \text{arccos}(z_2), \\
    \end{aligned}
\right])
\end{align}

In the local matching phase, each goal object region $o_j$ and its corresponding candidate object regions are processed to generate object correspondences, denoted as $\mathcal{M}$. These candidates are evaluated sequentially. For our 2D-2D matching, it is essential to consider the diversity of objects and the potential variability of environments. We employ the transformer-based feature matching network, LoFTR\cite{sun_loftr_2021} for this matching step due to its reliable performance. Using Eq. \ref{2D_match}, the 2D-2D correspondences $\mathcal{M}_{2D}$ are established between the goal image segmentation $S_j$ and candidate image segmentation $S_{cand}$, as shown in Fig. \ref{pipeline}(f). Following this, 2D-3D correspondences $\mathcal{M}_{3D}$ can be derived utilizing the point cloud $P_{cand}$ of the candidate object. Finally, the relative pose $T_{j}$ between goal scene $s^*$ and initial scene $s$ is computed by solving the PnP problem with $\mathcal{M}_{3D}$, as shown in Fig. \ref{pipeline}(g).
\begin{equation}
\label{2D_match}
\mathcal{M}_{2D} = \text{LoFTR}(\text{resize}(S_{cand}), \text{resize}(S_j))
\end{equation}

\begin{algorithm}[H]
\caption{Pseudocode of the Planning Module}
\label{planning_module}
\begin{algorithmic}[1]
\STATE \textbf{Input:} Object pose offsets $\left\{T_1, \ldots, T_{K}\right\}$
\STATE Initialize RemainingObjects = $[object_1, \dots, object_K]$
\STATE Initialize FailureCnts = $[cnt_1, \dots, cnt_K] = 0$, 
\WHILE{True}
    \FOR{$i = 1$ to size(RemainingObjects)}
        \STATE Move to the home viewpoint
        \STATE Correct $T_i$ based on the current observation
        \STATE Collision = (check if applying $T_i$ to $object_i$ is likely to collide others)
        \IF{not Collision}
            \STATE Move $object_i$ with $T_i$
            \STATE Remove $object_i$ from RemainingObjects
        \ELSE
            \STATE $cnt_i$ += 1
            \IF{$cnt_i$ $>$ Thres}
                \STATE Move $object_i$ to a random collision-free space
            \ENDIF
        \ENDIF
    \ENDFOR
    \IF{size(RemainingObjects) = 0 or TryCnt $>$ Thres}
        \STATE break loop
    \ENDIF
\ENDWHILE
\end{algorithmic}
\end{algorithm}

\begin{table*}[t]
    \centering
    \caption{Results of pose estimation accuracy}
    \label{PoseEstimation}
    \begin{tabular}{P{2.8cm}P{2.4cm}P{2.4cm}P{0.4cm}P{2.4cm}P{2.4cm}}
    \toprule 
    & \multicolumn{2}{c}{ Rot. $\in\left[-60^{\circ}, 60^{\circ}\right]$} & & \multicolumn{2}{c}{ Rot. $\in\left[-180^{\circ}, 180^{\circ}\right]$} \\
    \cmidrule(r){2-3} \cmidrule(r){5-6} 
    Method & $\begin{array}{c}\text { Median } \\
            |\Delta \theta|\left(\text { in }^{\circ}\right)\end{array}$ & $\begin{array}{c}\text { Median } \\
            |\Delta t| \text { (in cm) }\end{array}$ & & $\begin{array}{c}\text { Median } \\
            |\Delta \theta|\left(\text { in }^{\circ}\right)\end{array}$ & $\begin{array}{c}\text { Median } \\
            |\Delta t| \text { (in cm) }\end{array}$ \\
    \midrule 
    IFOR Baseline                 & 3.60 & 1.20 & & 13.70 & 2.70 \\
    Ours + Single-view   & 7.21 & 2.24 & & 153.67 & 13.10 \\
    Ours + Multi-view    & $\mathbf{0.17}$ & $\mathbf{0.17}$ & & $\mathbf{0.25}$ & $\mathbf{0.18}$ \\
    \bottomrule
    \end{tabular}
    \vspace*{-0.12in}
\end{table*}

\begin{table}[t]
    \centering
    \caption{Results of task completion}
    \label{SceneCompeletion}
    \begin{tabular}{P{1.8cm}P{2.4cm}P{2.7cm}}
    \toprule
    Setting  &  Method  &  $\begin{array}{c}\text { Task Completion } \\
            \text { (in \%) }\end{array}$ \\ \midrule
    \multirow{2}{*}{Multi-step} &  IFOR Baseline & 59.7\%    \\ 
    &  Ours Multi-view & \textbf{\underline{64.0\%}} \\
    \midrule
    \multirow{2}{*}{One-step}        &  Ours + Single-view & 10.0\%  \\
                                     &  Ours + Multi-view  & \underline{56.7\%}\\
                                     \bottomrule
    \end{tabular}
    \vspace*{-0.15in}
\end{table}

\subsection{Planning Module}
Given the poses $\left\{T_j\right\}_{j=1}^K$ of objects, the planning module directs the robot's interaction with the environment to complete the rearrangement task. The pseudo-code for the planning module is shown in Algorithm \ref{planning_module}. We utilize Contact-Graspnet\cite{sundermeyer_contact-graspnet_2021} to determine the grasping pose of objects. The intended placement pose of an object is calculated by applying $T_j$ to its grasping pose. Each object can be rearranged by picking it up from its grasping pose and placing it in the intended placement pose. Meanwhile, collision detection is managed by MoveIt!\cite{chitta_moveit_2012}.

A strategy involving multiple iterative attempts is executed, as shown in lines 4-5 of Algorithm \ref{planning_module}. Objects are processed in a predefined order. For each object, its associated pose $T$ is utilized for collision detection. If the object can be moved without causing any collisions, it is rearranged and then excluded from the following iterations (lines 8-11). If a collision occurs, the failed attempt for that object is logged. Once a preset threshold of failures for a specific object is reached, it is moved to a random collision-free position (lines 13-15). An iteration concludes when all objects have been successfully rearranged. 

\begin{figure*}[t]
    \centerline{\includegraphics[width=1.0\textwidth, clip=true,trim=0.15in 0.1in 0.1in 0in]{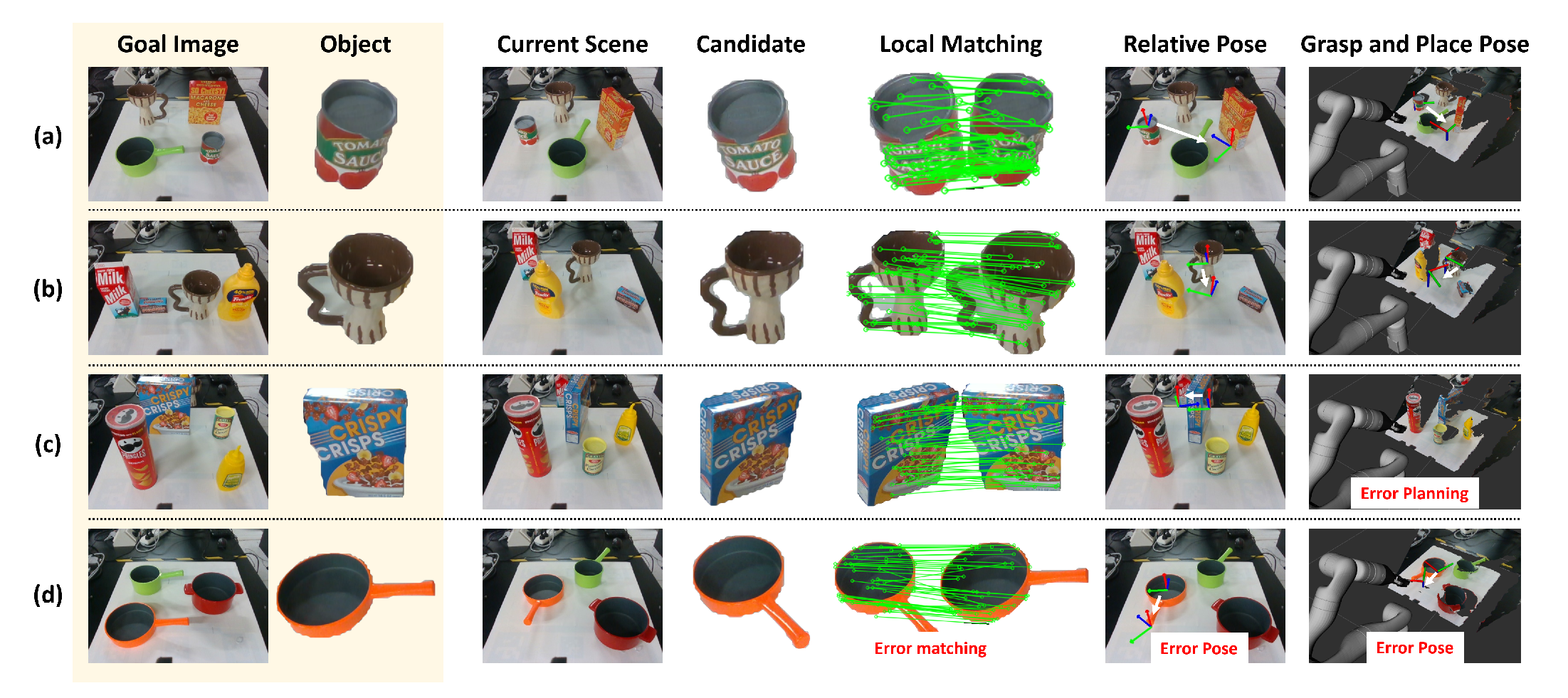}}
    \caption{Qualitative results. Each row is a visualization of the intermediate results for an object in an experimental scene. The candidate is the top retrieved object region within the database. In the figure of the relative pose, the arrow points from the initial object coordinate system of the current scene to the estimated object pose of the goal image. The last column shows the feasible grasping pose and its corresponding placement pose for the current object.}
    \label{qua_exp}
    \vspace*{-0.15in}
\end{figure*}

\section{EXPERIMENTS}
\subsection{Quantitative Experiments}
For quantitative evaluation, we conduct experiments on a synthetic dataset. This dataset is parallel to the one used by the current state-of-the-art method, IFOR\cite{goyal_ifor_2022} Baseline. We analyzed the superiority of the proposed method through side-by-side comparisons. 


\begin{figure}[H]
    \centerline{\includegraphics[width=0.5\textwidth, clip=true,trim=0.0in 0.0in 0.0in 0.1in]{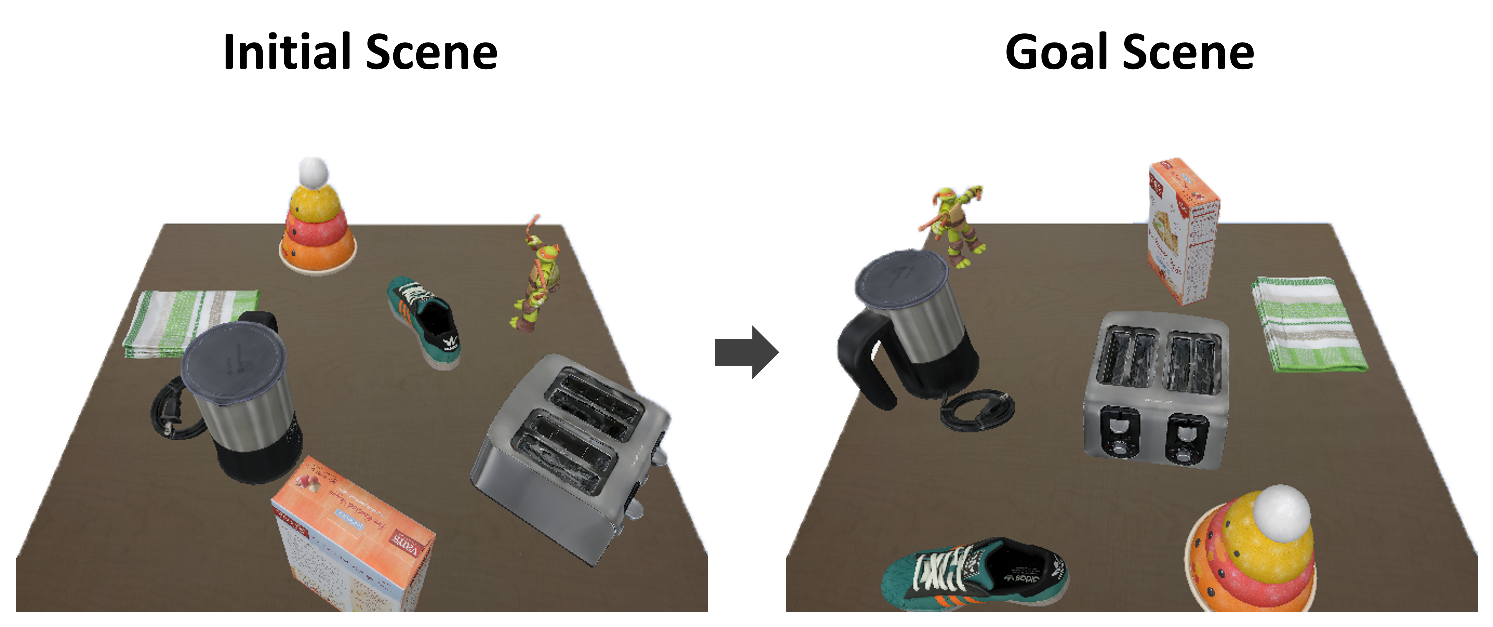}}
    \caption{Visualizations of an initial scene and its corresponding goal scene within the generated dataset.}
    \label{dateset_example}
    \vspace*{-0.15in}
\end{figure}

\noindent\textbf{Dataset.} We assembled our dataset in the Pybullet simulation environment\cite{coumans_pybullet_2016}. The synthetic collection comprises 150 tabletop scenes, each containing between 1 to 9 random objects. These objects are sourced from the Google Scanned Objects dataset\cite{downs_google_2022}, which features 115 unique objects distributed across 7 categories. Each category encompasses at least 10 distinct objects. For dataset creation, we initially select a random number of objects (ranging from 1 to 9) from the available 115 instances. An example is shown in Fig. \ref{dateset_example}. These selected objects are then arbitrarily positioned within a designated tabletop area and subjected to random planar rotations. The goal image is captured from a fixed frontal home pose. Subsequently, each object is shifted through a collision-free rotation and planar movement. Finally, the initial images are captured from a series of predefined viewpoints encircling the center of the tabletop.

\vspace{0.15\baselineskip}
\noindent\textbf{Baselines.} First, we compare the proposed multi-view method with the IFOR Baseline. The results for IFOR Baseline are sourced from its original paper, as its codebases are not publicly available. To make the comparison as fair as possible, our dataset was generated following the instructions given in its paper. Second, we compare our method using multi-view observation and using single-view observation (our single-view baseline). Our single-view baseline constructs the database with a single frame. 

\vspace{0.15\baselineskip}

\noindent\textbf{Pose estimation.} With accurate pose estimation, a rearrangement system can reduce redundant manipulations used for adjusting the object's pose, which improves its efficiency. To evaluate the accuracy of the object pose estimation, we employ the metric measuring the median translation and rotation errors across all objects\cite{goyal_ifor_2022, tang_selective_nodate}. Due to the difference in the object's angle between the goal scene and current scene largely determines the difficulty of pose estimation, we evaluate pose estimation accuracy under two distinct scenarios: a) objects with minor rotation, ranging between -60$^{\circ}$ and 60$^{\circ}$ and b) objects with full rotation, spanning from -180$^{\circ}$ to 180$^{\circ}$.

Experimental results in Table \ref{PoseEstimation} show that the IFOR Baseline and our single-view baseline in the case of full rotation, are markedly lower compared to the cases involving only minor rotations. This decline in performance can be attributed to its reliance on single-view information. Conversely, the accuracy levels exhibited by our proposed multi-view method are not only superior to the baseline but also consistent across both rotation scenarios. While IFOR Baseline is trained on an expansive dataset of around 54,000 rearrangement scenes from both ShapeNet\cite{chang_shapenet_2015} and the Google Scanned Dataset\cite{downs_google_2022}, our multi-view method consistently outperforms it for most objects. This improved performance is achieved without training, highlighting the effectiveness of using multi-view observations. 

\vspace{0.15\baselineskip}

\noindent\textbf{Task completion.} To further verify that accurate pose estimation improves the system's efficiency, we evaluate the system's efficiency using the task completion rate metric\cite{goyal_ifor_2022, tang_selective_nodate} in two distinct scenarios: a) one-step setting, where each object is subjected to a single manipulation, excluding the mandatory move to the free buffer; and b) multi-step setting, where no restriction on the number of manipulations is imposed. In general, the system works more effectively in a one-step setting than in a multi-step setting. A scene is considered successfully rearranged when both the position and rotation errors for all its objects fall below thresholds of $2\ cm$ and $5^{\circ}$.

Experimental results in Table \ref{SceneCompeletion} show a significant enhancement in our proposed multi-view system compared to the IFOR Baseline and our single-view baseline under the multi-step setting. This improvement underscores that our system is nearing its theoretical performance limit. In the one-step setting, our proposed method exhibits competitive performance against the multi-step baseline. In contrast, the IFOR Baseline requires an average of 2-3 manipulations for each object. This suggests that our proposed system enhances efficiency by performing complete multi-view observations of the scene before manipulating the objects.

\subsection{Qualitative Experiments}
To evaluate the practicality and applicability of our proposed method, we conduct qualitative evaluation through physical experiments. Our experimental setup employs a Kinova Gen3 robot, augmented with a wrist-mounted RealSense D435i camera. The experiments encompass five unique scenes featuring 20 novel objects, with each scene presenting a mix of 2 to 5 objects. Some qualitative results are provided in Fig. \ref{qua_exp}. The videos of the experiments are provided in the Supplementary Materials.

Physical experiments confirm two major challenges hindering successful object rearrangement. Firstly, inaccuracies in the perception module result in incorrect pose predictions, as shown in Fig. \ref{qua_exp}(d). This happens because the local matching module sometimes produces incorrect correspondences in complex physical environments. A robust local matching module for object region matching can alleviate this problem. Secondly, the robot arm's motion planning module struggles to generate executable collision-free trajectories, as shown in Fig. \ref{qua_exp}(c). We believe that this problem can be addressed by integrating the TAMP framework, which focuses on the feasibility of planning.

\section{CONCLUSIONS}
We propose the first multi-view object rearrangement system tailored for the image-goal setting. Leveraging multi-view RGB-D observations of the current scene, we construct a hierarchical database with each object region from the observations serving as a database item. For every object in the goal RGB image, we calculate its pose through retrieval and local matching steps based on the database. Experimental results show that our method performs more accurate pose estimation compared with single-view methods, which reduces redundant manipulations and improves the efficiency of the system.

At present, we gather multi-view observations from predefined viewpoints. In our upcoming work, we intend to implement an active observation strategy. Furthermore, we plan to introduce an online learning methodology for the feature descriptor extractor tailored to the object-level scenario. 



\bibliographystyle{IEEEtran}
\balance
\bibliography{main}

\begin{thebibliography}{10}
\providecommand{\url}[1]{#1}
\csname url@samestyle\endcsname
\providecommand{\newblock}{\relax}
\providecommand{\bibinfo}[2]{#2}
\providecommand{\BIBentrySTDinterwordspacing}{\spaceskip=0pt\relax}
\providecommand{\BIBentryALTinterwordstretchfactor}{4}
\providecommand{\BIBentryALTinterwordspacing}{\spaceskip=\fontdimen2\font plus
\BIBentryALTinterwordstretchfactor\fontdimen3\font minus \fontdimen4\font\relax}
\providecommand{\BIBforeignlanguage}[2]{{%
\expandafter\ifx\csname l@#1\endcsname\relax
\typeout{** WARNING: IEEEtran.bst: No hyphenation pattern has been}%
\typeout{** loaded for the language `#1'. Using the pattern for}%
\typeout{** the default language instead.}%
\else
\language=\csname l@#1\endcsname
\fi
#2}}
\providecommand{\BIBdecl}{\relax}
\BIBdecl

\bibitem{batra_rearrangement_2020}
\BIBentryALTinterwordspacing
D.~Batra, A.~X. Chang, S.~Chernova, A.~J. Davison, J.~Deng, V.~Koltun, S.~Levine, J.~Malik, I.~Mordatch, R.~Mottaghi, M.~Savva, and H.~Su, ``\BIBforeignlanguage{en}{Rearrangement: {A} {Challenge} for {Embodied} {AI}},'' Nov. 2020, number: arXiv:2011.01975 arXiv:2011.01975 [cs]. [Online]. Available: \url{http://arxiv.org/abs/2011.01975}
\BIBentrySTDinterwordspacing

\bibitem{qureshi_nerp_2021}
\BIBentryALTinterwordspacing
A.~Qureshi, A.~Mousavian, C.~Paxton, M.~Yip, and D.~Fox, ``\BIBforeignlanguage{en}{{NeRP}: {Neural} {Rearrangement} {Planning} for {Unknown} {Objects}},'' in \emph{\BIBforeignlanguage{en}{Robotics: {Science} and {Systems} {XVII}}}.\hskip 1em plus 0.5em minus 0.4em\relax Robotics: Science and Systems Foundation, Jul. 2021. [Online]. Available: \url{http://www.roboticsproceedings.org/rss17/p072.pdf}
\BIBentrySTDinterwordspacing

\bibitem{tang_selective_nodate}
B.~Tang and G.~S. Sukhatme, ``Selective {Object} {Rearrangement} in {Clutter},'' in \emph{6th {Annual} {Conference} on {Robot} {Learning}}, ser. {CoRL2022}.

\bibitem{goyal_ifor_2022}
A.~Goyal, A.~Mousavian, C.~Paxton, Y.-W. Chao, B.~Okorn, J.~Deng, and D.~Fox, ``Ifor: {Iterative} flow minimization for robotic object rearrangement,'' in \emph{Proceedings of the {IEEE}/{CVF} {Conference} on {Computer} {Vision} and {Pattern} {Recognition}}, 2022, pp. 14\,787--14\,797.

\bibitem{lepetit_epnp_2009}
V.~Lepetit, F.~Moreno-Noguer, and P.~Fua, ``{EPnP}: {An} accurate {O}(n) solution to the {PnP} problem,'' \emph{International journal of computer vision}, vol.~81, pp. 155--166, 2009, publisher: Springer.

\bibitem{garrett_online_2020}
C.~R. Garrett, C.~Paxton, T.~Lozano-Pérez, L.~P. Kaelbling, and D.~Fox, ``Online replanning in belief space for partially observable task and motion problems,'' in \emph{2020 {IEEE} {International} {Conference} on {Robotics} and {Automation} ({ICRA})}.\hskip 1em plus 0.5em minus 0.4em\relax IEEE, 2020, pp. 5678--5684.

\bibitem{krontiris_efficiently_2016}
\BIBentryALTinterwordspacing
A.~Krontiris and K.~E. Bekris, ``\BIBforeignlanguage{en}{Efficiently solving general rearrangement tasks: {A} fast extension primitive for an incremental sampling-based planner},'' in \emph{\BIBforeignlanguage{en}{2016 {IEEE} {International} {Conference} on {Robotics} and {Automation} ({ICRA})}}.\hskip 1em plus 0.5em minus 0.4em\relax Stockholm, Sweden: IEEE, May 2016, pp. 3924--3931. [Online]. Available: \url{http://ieeexplore.ieee.org/document/7487581/}
\BIBentrySTDinterwordspacing

\bibitem{krontiris_dealing_2015}
------, ``Dealing with {Difficult} {Instances} of {Object} {Rearrangement}.'' in \emph{Robotics: {Science} and {Systems}}, vol. 1123, 2015.

\bibitem{wang_uniform_2021}
\BIBentryALTinterwordspacing
R.~Wang, K.~Gao, D.~Nakhimovich, J.~Yu, and K.~E. Bekris, ``\BIBforeignlanguage{en}{Uniform {Object} {Rearrangement}: {From} {Complete} {Monotone} {Primitives} to {Efficient} {Non}-{Monotone} {Informed} {Search}},'' in \emph{\BIBforeignlanguage{en}{2021 {IEEE} {International} {Conference} on {Robotics} and {Automation} ({ICRA})}}.\hskip 1em plus 0.5em minus 0.4em\relax Xi'an, China: IEEE, May 2021, pp. 6621--6627. [Online]. Available: \url{https://ieeexplore.ieee.org/document/9561716/}
\BIBentrySTDinterwordspacing

\bibitem{labbe_monte-carlo_2020}
Y.~Labbé, S.~Zagoruyko, I.~Kalevatykh, I.~Laptev, J.~Carpentier, M.~Aubry, and J.~Sivic, ``Monte-carlo tree search for efficient visually guided rearrangement planning,'' \emph{IEEE Robotics and Automation Letters}, vol.~5, no.~2, pp. 3715--3722, 2020, publisher: IEEE.

\bibitem{wang_efficient_2022}
R.~Wang, Y.~Miao, and K.~E. Bekris, ``Efficient and high-quality prehensile rearrangement in cluttered and confined spaces,'' in \emph{2022 {International} {Conference} on {Robotics} and {Automation} ({ICRA})}.\hskip 1em plus 0.5em minus 0.4em\relax IEEE, 2022, pp. 1968--1975.

\bibitem{wang_lazy_2022}
R.~Wang, K.~Gao, J.~Yu, and K.~Bekris, ``Lazy rearrangement planning in confined spaces,'' in \emph{Proceedings of the {International} {Conference} on {Automated} {Planning} and {Scheduling}}, vol.~32, 2022, pp. 385--393.

\bibitem{liu_structformer_2022}
W.~Liu, C.~Paxton, T.~Hermans, and D.~Fox, ``Structformer: {Learning} spatial structure for language-guided semantic rearrangement of novel objects,'' in \emph{2022 {International} {Conference} on {Robotics} and {Automation} ({ICRA})}.\hskip 1em plus 0.5em minus 0.4em\relax IEEE, 2022, pp. 6322--6329.

\bibitem{liu_structdiffusion_2022}
\BIBentryALTinterwordspacing
W.~Liu, T.~Hermans, S.~Chernova, and C.~Paxton, ``{StructDiffusion}: {Object}-{Centric} {Diffusion} for {Semantic} {Rearrangement} of {Novel} {Objects},'' in \emph{Workshop on {Language} and {Robotics} at {CoRL} 2022}, 2022. [Online]. Available: \url{https://openreview.net/forum?id=pPGE9AyvukF}
\BIBentrySTDinterwordspacing

\bibitem{kapelyukh_dall-e-bot_2023}
I.~Kapelyukh, V.~Vosylius, and E.~Johns, ``Dall-e-bot: {Introducing} web-scale diffusion models to robotics,'' \emph{IEEE Robotics and Automation Letters}, 2023, publisher: IEEE.

\bibitem{sundermeyer_contact-graspnet_2021}
M.~Sundermeyer, A.~Mousavian, R.~Triebel, and D.~Fox, ``Contact-graspnet: {Efficient} 6-dof grasp generation in cluttered scenes,'' in \emph{2021 {IEEE} {International} {Conference} on {Robotics} and {Automation} ({ICRA})}.\hskip 1em plus 0.5em minus 0.4em\relax IEEE, 2021, pp. 13\,438--13\,444.

\bibitem{tang_task-oriented_2023}
\BIBentryALTinterwordspacing
C.~Tang, D.~Huang, L.~Meng, W.~Liu, and H.~Zhang, ``\BIBforeignlanguage{en}{Task-{Oriented} {Grasp} {Prediction} with {Visual}-{Language} {Inputs}},'' Feb. 2023, arXiv:2302.14355 [cs]. [Online]. Available: \url{http://arxiv.org/abs/2302.14355}
\BIBentrySTDinterwordspacing

\bibitem{tang_graspgpt_2023}
\BIBentryALTinterwordspacing
C.~Tang, D.~Huang, W.~Ge, W.~Liu, and H.~Zhang, ``\BIBforeignlanguage{en}{{GraspGPT}: {Leveraging} {Semantic} {Knowledge} from a {Large} {Language} {Model} for {Task}-{Oriented} {Grasping}},'' Jul. 2023, arXiv:2307.13204 [cs]. [Online]. Available: \url{http://arxiv.org/abs/2307.13204}
\BIBentrySTDinterwordspacing

\bibitem{danielczuk_object_2021}
M.~Danielczuk, A.~Mousavian, C.~Eppner, and D.~Fox, ``Object rearrangement using learned implicit collision functions,'' in \emph{2021 {IEEE} {International} {Conference} on {Robotics} and {Automation} ({ICRA})}.\hskip 1em plus 0.5em minus 0.4em\relax IEEE, 2021, pp. 6010--6017.

\bibitem{segal_generalized-icp_2009}
A.~Segal, D.~Haehnel, and S.~Thrun, ``Generalized-icp.'' in \emph{Robotics: science and systems}, vol.~2.\hskip 1em plus 0.5em minus 0.4em\relax Seattle, WA, 2009, p. 435, issue: 4.

\bibitem{castro_posematcher_2023}
P.~Castro and T.-K. Kim, ``{PoseMatcher}: {One}-shot {6D} {Object} {Pose} {Estimation} by {Deep} {Feature} {Matching},'' \emph{arXiv preprint arXiv:2304.01382}, 2023.

\bibitem{he_onepose_2022}
X.~He, J.~Sun, Y.~Wang, D.~Huang, H.~Bao, and X.~Zhou, ``Onepose++: {Keypoint}-free one-shot object pose estimation without {CAD} models,'' \emph{Advances in Neural Information Processing Systems}, vol.~35, pp. 35\,103--35\,115, 2022.

\bibitem{sun_onepose_2022}
J.~Sun, Z.~Wang, S.~Zhang, X.~He, H.~Zhao, G.~Zhang, and X.~Zhou, ``Onepose: {One}-shot object pose estimation without cad models,'' in \emph{Proceedings of the {IEEE}/{CVF} {Conference} on {Computer} {Vision} and {Pattern} {Recognition}}, 2022, pp. 6825--6834.

\bibitem{liu_gen6d_2022}
Y.~Liu, Y.~Wen, S.~Peng, C.~Lin, X.~Long, T.~Komura, and W.~Wang, ``{Gen6D}: {Generalizable} model-free 6-{DoF} object pose estimation from {RGB} images,'' in \emph{European {Conference} on {Computer} {Vision}}.\hskip 1em plus 0.5em minus 0.4em\relax Springer, 2022, pp. 298--315.

\bibitem{wen_bundlesdf_2023}
B.~Wen, J.~Tremblay, V.~Blukis, S.~Tyree, T.~Müller, A.~Evans, D.~Fox, J.~Kautz, and S.~Birchfield, ``{BundleSDF}: {Neural} 6-{DoF} {Tracking} and {3D} {Reconstruction} of {Unknown} {Objects},'' in \emph{Proceedings of the {IEEE}/{CVF} {Conference} on {Computer} {Vision} and {Pattern} {Recognition}}, 2023, pp. 606--617.

\bibitem{shugurov_osop_2022}
I.~Shugurov, F.~Li, B.~Busam, and S.~Ilic, ``Osop: {A} multi-stage one shot object pose estimation framework,'' in \emph{Proceedings of the {IEEE}/{CVF} {Conference} on {Computer} {Vision} and {Pattern} {Recognition}}, 2022, pp. 6835--6844.

\bibitem{liu_grounding_2023}
S.~Liu, Z.~Zeng, T.~Ren, F.~Li, H.~Zhang, J.~Yang, C.~Li, J.~Yang, H.~Su, J.~Zhu, and {others}, ``Grounding dino: {Marrying} dino with grounded pre-training for open-set object detection,'' \emph{arXiv preprint arXiv:2303.05499}, 2023.

\bibitem{kirillov_segment_2023}
A.~Kirillov, E.~Mintun, N.~Ravi, H.~Mao, C.~Rolland, L.~Gustafson, T.~Xiao, S.~Whitehead, A.~C. Berg, W.-Y. Lo, P.~Dollár, and R.~Girshick, ``Segment {Anything},'' \emph{arXiv:2304.02643}, 2023.

\bibitem{ali-bey_mixvpr_2023}
A.~Ali-Bey, B.~Chaib-Draa, and P.~Giguere, ``Mixvpr: {Feature} mixing for visual place recognition,'' in \emph{Proceedings of the {IEEE}/{CVF} {Winter} {Conference} on {Applications} of {Computer} {Vision}}, 2023, pp. 2998--3007.

\bibitem{sun_loftr_2021}
J.~Sun, Z.~Shen, Y.~Wang, H.~Bao, and X.~Zhou, ``{LoFTR}: {Detector}-free local feature matching with transformers,'' in \emph{Proceedings of the {IEEE}/{CVF} conference on computer vision and pattern recognition}, 2021, pp. 8922--8931.

\bibitem{chitta_moveit_2012}
\BIBentryALTinterwordspacing
S.~Chitta, I.~Sucan, and S.~Cousins, ``\BIBforeignlanguage{en}{{MoveIt}! [{ROS} {Topics}]},'' \emph{\BIBforeignlanguage{en}{IEEE Robotics \& Automation Magazine}}, vol.~19, no.~1, pp. 18--19, Mar. 2012. [Online]. Available: \url{http://ieeexplore.ieee.org/document/6174325/}
\BIBentrySTDinterwordspacing

\bibitem{coumans_pybullet_2016}
\BIBentryALTinterwordspacing
E.~Coumans and Y.~Bai, ``{PyBullet}, a {Python} module for physics simulation for games, robotics and machine learning,'' 2016. [Online]. Available: \url{http://pybullet.org}
\BIBentrySTDinterwordspacing

\bibitem{downs_google_2022}
L.~Downs, A.~Francis, N.~Koenig, B.~Kinman, R.~Hickman, K.~Reymann, T.~B. McHugh, and V.~Vanhoucke, ``Google scanned objects: {A} high-quality dataset of 3d scanned household items,'' in \emph{2022 {International} {Conference} on {Robotics} and {Automation} ({ICRA})}.\hskip 1em plus 0.5em minus 0.4em\relax IEEE, 2022, pp. 2553--2560.

\bibitem{chang_shapenet_2015}
A.~X. Chang, T.~Funkhouser, L.~Guibas, P.~Hanrahan, Q.~Huang, Z.~Li, S.~Savarese, M.~Savva, S.~Song, H.~Su, J.~Xiao, L.~Yi, and F.~Yu, ``{ShapeNet}: {An} {Information}-{Rich} {3D} {Model} {Repository},'' Stanford University — Princeton University — Toyota Technological Institute at Chicago, Tech. Rep. arXiv:1512.03012 [cs.GR], 2015.

\end{thebibliography}

\end{document}